\documentclass[11pt,letterpaper]{arxiv}
\pdfoutput=1
\usepackage[numbers,sort&compress]{natbib}
\usepackage{parskip}
\usepackage{arxiv}
\usepackage{tikz}
\usetikzlibrary{arrows.meta}

\definecolor{darkblue}{HTML}{234A6F}

\newcommand{\web}{https://yingyuan0414.github.io/fomo-dex/}

\title{Force-Modulated Dexterity: Unified Compliance for Contact-Rich Manipulation}
\author[1,2]{Ying Yuan}
\author[2]{David Held}
\author[1]{Jitendra Malik}
\author[1]{Haozhi Qi}
\affil[1]{Amazon FAR}
\affil[2]{Carnegie Mellon University}
\projectwebsite{\web}{\texttt{\web}}

\renewcommand{\preabstractcontent}{%
\begin{center}
\includegraphics[width=\linewidth]{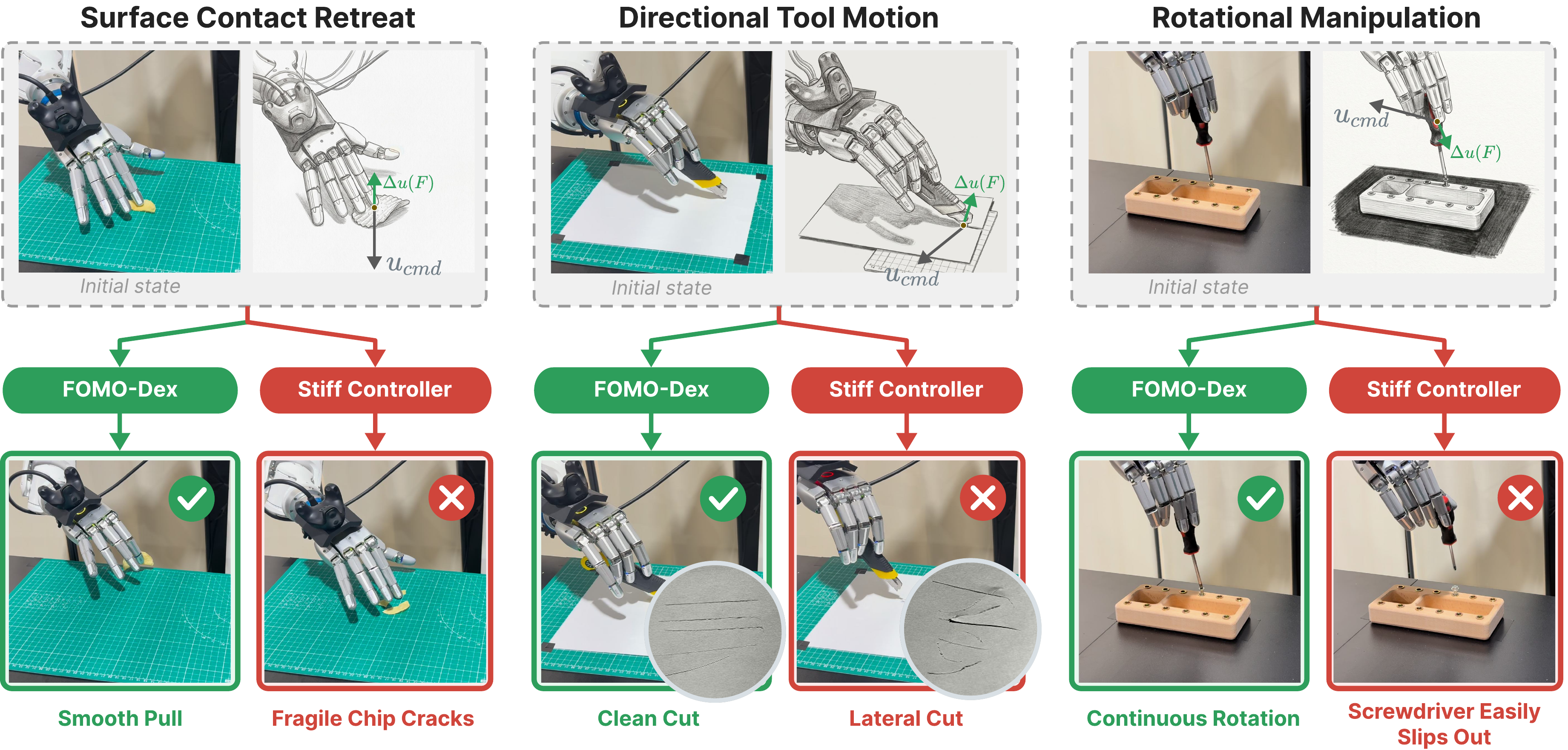}

\captionof{figure}{\textit{\underline{Fo}rce-\underline{Mo}dulated \underline{Dex}terity (FOMO-Dex)} uses force feedback to adjust selected components of a motion command while leaving the others unchanged. The sketches show nominal commands (gray) and force-driven corrections (green). The same principle relieves pressure on a fragile object (left), reduces lateral loading during cutting (middle), and regulates grip during screwdriving (right). The outcomes illustrate how these corrections avoid the object damage and loss of tool control seen with stiff control.}

\label{fig:teaser-draft}

\end{center}
}

\begin{document}

\begin{abstract}
Teleoperation is widely used to collect high-quality demonstrations for robot
learning. Dexterous manipulation, however, requires coordinated control of
motion and contact forces. Operators can specify motion, but struggle to
reproduce the subtle force adjustments of human manipulation without feeling
the robot's contacts. We introduce \emph{Force-Modulated Dexterity}
(FOMO-Dex), a unified compliance formulation for contact-rich manipulation.
FOMO-Dex expresses nominal actions in coordinates defined by task geometry and
contact structure. It applies force-driven motion corrections along selected
directions while leaving the remaining command components unchanged.
Across card picking, paper cutting, screwdriving, and nut-and-bolt
assembly, FOMO-Dex reduces excessive contact forces compared with stiff
control. A user study shows that operators collect more successful
demonstration data per hour after brief practice.
For policy learning, imitating the corrected actions improves card-picking
success over imitating raw operator commands. Policies trained on FOMO-Dex
demonstrations also achieve more rotation in nut-and-bolt manipulation and
screwdriving than those trained on demonstrations collected with the stiff
controller.
\end{abstract}

\maketitle

\section{Introduction}

Realizing human-like dexterity has long been a central goal of robotics. Recent
data-driven approaches offer a promising path toward this goal, but their
progress depends on collecting high-quality demonstrations. Teleoperation
provides a direct way to collect such data~\cite{handa2020dexpilot,qin2023anyteleop,yang2024ace},
yet collecting reliable contact-rich demonstrations remains difficult.
Human manipulation relies on subtle sensorimotor
adjustments that are largely automatic and difficult to command explicitly. In
teleoperation, the operator cannot easily reproduce these adjustments without
directly feeling the robot's contacts. The resulting errors become especially
consequential during contact, where small inaccuracies can produce large and
inconsistent forces. Maintaining a stable force or deliberately changing it as
the task requires is therefore difficult through motion commands alone.

Compliance provides a natural way to accommodate uncertainty during physical
interaction~\cite{raibert1981hybrid,hogan1984impedance,
Bruyninckx1996SpecificationOF}. It has been used to regulate object motion
and grasp forces in multi-fingered hands~\cite{li2014object,pfanne2020object}
and to learn contact-rich manipulation skills~\cite{hou2025adaptive,choi2026wild}.
Applying the same idea to dexterous manipulation, however,
is not straightforward. A multi-fingered hand can simultaneously make and
break several contacts. These contacts impose different motion constraints
and are mechanically coupled through the manipulated object. For a tool, the
robot should preserve motion along the task direction while responding to
forces in the constrained directions. For a grasp, it should regulate
fingertip forces without suppressing the motion that turns the object.
Simply making the arm and hand more compliant does not specify how to achieve
both. We must determine which action components should respond to contact
and which should remain unchanged.

We propose \emph{Force-Modulated Dexterity} (FOMO-Dex), which treats the
commanded motion as a nominal action and uses force feedback to modify
selected components. We define interaction directions from the task geometry
and current contacts, then convert force errors along these directions into
motion corrections. Directions excluded from modulation retain the nominal
command. The corrections can act through either the arm or the fingers,
using wrist force-torque, fingertip tactile, or joint-torque measurements as
appropriate for the contact. Because the formulation operates on motion
commands, the nominal action can come from either a teleoperator or a learned
policy.

The same principle takes a different form in each task. To pick up a thin card,
the contacting finger retreats when its load against the supporting surface
becomes excessive. During cutting, a utility knife yields to lateral forces
while retaining the commanded motion along the cut. In screwdriving and
nut-and-bolt manipulation, the controller regulates fingertip forces while
leaving the commanded rotation unchanged. The force targets and feedback laws
differ, but each realization follows the same principle: task and contact
geometry determine where force feedback may change the motion.

Across these four tasks, FOMO-Dex reduces excessive contact forces compared
with stiff control. In screwdriving, it also increases mean rotation from
4.2 to 34.7 turns per episode while reducing tool slips
(Table~\ref{tab:controller_comparison}). A user study further shows that
operators collect more successful demonstration data per hour with FOMO-Dex
after brief practice (Table~\ref{tab:user_study}). These results evaluate
compliance not only by the forces it reduces, but also by the manipulation
and demonstration collection it enables.

We also study how these demonstrations should be used for policy learning.
During teleoperation, we record both the nominal commands and the
force-modulated actions sent to the robot. In card picking, directly
imitating the corrected actions produces higher lift success than learning
raw commands and applying force modulation at deployment. For nut-and-bolt
manipulation and screwdriving, policies trained on FOMO-Dex demonstrations
achieve more rotation than policies trained on stiff-control demonstrations
(Table~\ref{tab:bc_results}). Force modulation thus supports both
demonstration collection and policy learning.

The contributions of this work are:
\begin{itemize}
    \item a unified force-modulation interface for dexterous manipulation that
    adjusts nominal arm and finger commands in directions defined by task
    geometry and contact structure;
    \item several realizations of this formulation for unilateral load relief,
    translational tool compliance, and multi-contact force regulation, demonstrated
    across four contact-rich dexterous tasks; and
    \item a teleoperation-to-learning approach that records paired nominal and
    force-modulated actions and uses the corrected actions to train autonomous
    visuomotor policies.
\end{itemize}

\section{Related Work}

\subsection{Dexterous Teleoperation}

Recent dexterous teleoperation systems map tracked human motion to high-DoF
robot hands~\cite{qin2022one,shaw2024bimanual}. These interfaces
provide an intuitive way to command dexterous motion, but motion tracking
alone does not regulate contact forces. Small errors can produce excessive
forces or unstable contact, and operators may struggle to correct them
without directly feeling the robot's interaction with the environment.
To help operators respond to contact, Bunny-VisionPro and DOGlove provide
haptic feedback~\cite{ding2025bunny,ZhangH-RSS-25}. DexUMI instead collects
demonstrations through a wearable hand exoskeleton, allowing the demonstrator
to interact directly with objects and feel their response~\cite{xu2025dexumi}.

DexScrew~\cite{hsieh2026dexscrew} uses simulation-trained finger-motion
primitives to assist teleoperation and collect demonstrations for tactile
policy learning. Robot-side compliance instead adjusts the response to contact without
requiring the operator to make every correction. Zeng et
al.~\cite{zeng2021learning} adapt joint impedance and feedforward forces
while tracking teleoperated hand postures. DexTeleop-0~\cite{liu2026dexteleop}
uses tactile measurements and contact locations to optimize joint-command
corrections that regulate fingertip loads and balance the object wrench.
FOMO-Dex shares this use of robot-side force feedback. Our focus is a common
formulation for fingertip loading and arm-mediated tool contact. Depending
on the interaction, selected directions either yield above a force limit
or regulate force toward a target, while the remaining command components
are unchanged.

\subsection{Compliance and Hybrid Position-Force Control}

Compliance has a long history in robotic manipulation. Hybrid position-force
control separates directions governed by motion and
force~\cite{raibert1981hybrid}, while impedance and admittance control specify
how motion and interaction forces are related~\cite{hogan1984impedance,
keemink2018admittance}. Task-frame formulations express motion and force
constraints in coordinates aligned with the interaction
geometry~\cite{Bruyninckx1996SpecificationOF}. Optimally-Conditioned Hybrid
Servoing chooses force- and velocity-controlled directions from a contact
model and a desired system motion to improve kinematic
conditioning~\cite{hou2021efficient}.

For multi-fingered hands, object-level impedance control already separates
grasp stability from object motion. Li et al.~\cite{li2014object} combine
grasp and manipulation impedance and learn the latter from demonstrations.
Pfanne et al.~\cite{pfanne2020object} regulate object pose while optimizing
internal forces under friction constraints, including during contact
reconfiguration. These methods control a desired object motion through the
fingers. FOMO-Dex instead takes nominal arm and finger commands and corrects
their response to contact without requiring an object-pose reference.

Minimalist Compliance Control~\cite{shi2026minimalist} makes task-space
admittance available on position-controlled robots by estimating external
wrenches from motor signals. It supports dexterous hands as well as arms
and can execute geometry-based compliance profiles. FOMO-Dex also builds
on established task-frame and admittance-control principles, rather than
introducing a new admittance law. We use them to specify task-dependent
force corrections during dexterous teleoperation and record the corrected
commands as demonstrations for policy learning.

\begin{figure}[t]
\centering
\includegraphics[width=\linewidth]{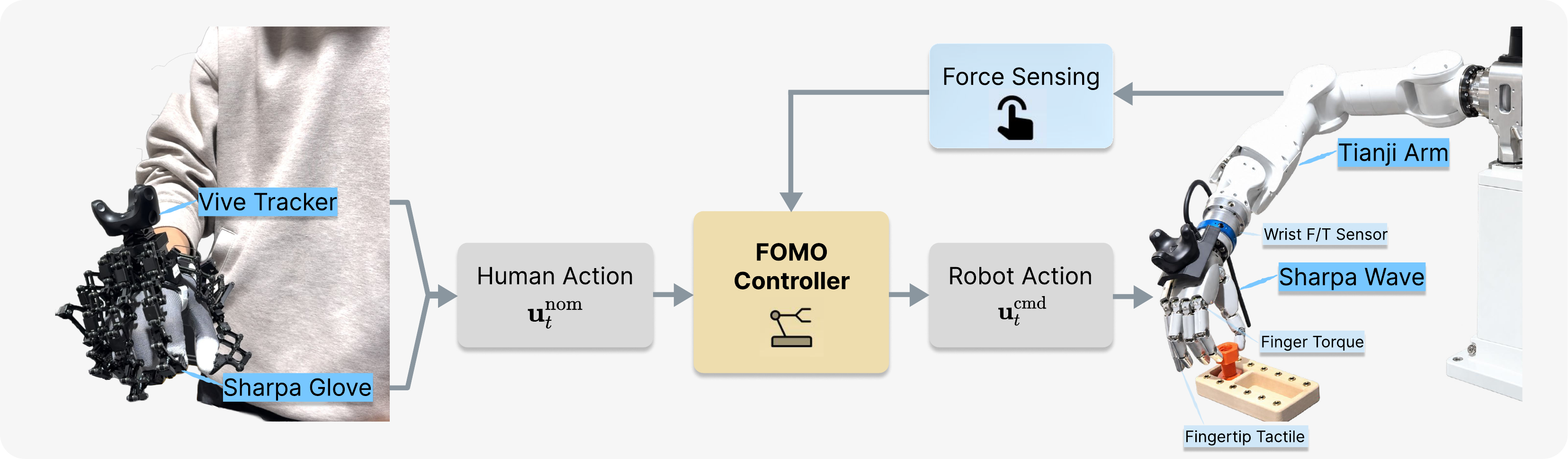}
\caption{System overview. A teleoperator or learned policy provides nominal
arm and finger commands. FOMO-Dex uses feedback from the wrist force-torque
sensor and the hand's tactile and joint-torque sensors to adjust selected
motion components before sending the commands to the robot.}
\label{fig:system-overview}
\end{figure}

\subsection{Learning Compliant Manipulation}

Earlier work on variable impedance learning uses human demonstrations to
estimate stiffness or impedance profiles and adapt them during
execution~\cite{abu2018force,abu2020variable,zhang2022learning,
zhou2025variable}. More recent diffusion policies incorporate force
measurements as observations for contact-rich
manipulation~\cite{shukla2025force,kang2025robotic}. Other approaches also
predict compliance parameters. Adaptive Compliance Policy (ACP) learns
spatially and temporally varying compliance by jointly predicting motion
targets and approximate stiffness~\cite{hou2025adaptive}. UMI-FT extends
this approach to in-the-wild multimodal demonstrations, predicting position
targets, grasp force, and stiffness~\cite{choi2026wild}. VIDP jointly predicts
robot motion and task-dependent compliance~\cite{khalil2026vidp}.

Beyond compliance gains, Force Policy~\cite{fang2026forcepolicy} learns an
interaction frame and motion/force selection for hybrid control from
demonstrations. Compliant Residual DAgger~\cite{xu2026compliant} uses a
compliant interface to collect human corrections and learns force-aware
residual policies. In FOMO-Dex, the task configuration specifies the
force-response directions and feedback laws. They act on nominal commands
without requiring a learned compliance or residual model.

Force information can also be obtained directly from human demonstrations.
OSMO~\cite{yin2026osmo} records normal and shear forces through a tactile
glove, while ForceBand~\cite{he2026forceband} estimates fingertip forces
from wrist sEMG. Both enrich human demonstrations with force information
for policy learning. FOMO-Dex instead uses robot-side force feedback to
correct actions during demonstration collection.

Force feedback also affects what actions a policy should imitate.
DexForce~\cite{chen2025dexforce} reconstructs fingertip position targets
from measured positions and contact forces in kinesthetic demonstrations,
then learns policies that output these force-informed targets.
Contact-Grounded Policy~\cite{xu2026contact} predicts robot-state and tactile
trajectories and learns a mapping from these predictions to executable
controller targets. We share the premise that measured motion alone does
not specify the action needed to reproduce contact. Our demonstrations
already contain both the nominal commands and the force-modulated commands
sent to the robot. We directly imitate the latter, rather than reconstructing
targets from measured states or learning a separate state-to-command mapping.

\section{Force-Modulated Dexterity}
\label{sec:method}
\subsection{Problem Formulation}
We consider a robotic arm and a multi-fingered hand teleoperated through
human motion tracking. The robot receives nominal arm and finger commands
together with force feedback from the wrist and hand. Directly executing
these commands during contact can produce excessive force, unstable tool
motion, or grasp loss. Our goal is to regulate contact forces toward
task-specific targets or limits while retaining the commanded motion in
directions that should remain stiff.

\subsection{System Setup}
Our system consists of a 7-DoF Tianji Marvin arm and a 22-DoF Sharpa Wave
dexterous hand (Fig.~\ref{fig:system-overview}). A six-axis force-torque
sensor at the wrist measures arm-level interaction forces and torques.
The hand provides fingertip tactile and joint-torque measurements for
finger-level feedback. A Sharpa Avatar glove tracks the operator's finger
motion, and an HTC Vive Tracker attached to the glove tracks the hand pose.

\subsection{Unified Force-Modulation Controller}

At time step $t$, the nominal command $\mathbf{u}^{\mathrm{nom}}_t$ consists
of a Cartesian end-effector pose and a vector of finger-joint targets.
The controller computes a force-derived correction $\Delta\mathbf{u}_t$
and combines it with the nominal command to obtain the command sent to the
robot,
\begin{equation}
    \mathbf{u}^{\mathrm{cmd}}_t
    =
    \mathbf{u}^{\mathrm{nom}}_t
    \oplus
    \Delta \mathbf{u}_t,
    \label{eq:addition}
\end{equation}
where $\oplus$ denotes addition for finger-joint commands and pose
composition for Cartesian end-effector commands. The operator provides the
nominal motion, and the controller adjusts its response to contact.

\begin{figure}[t]
\centering
\includegraphics[width=\linewidth]{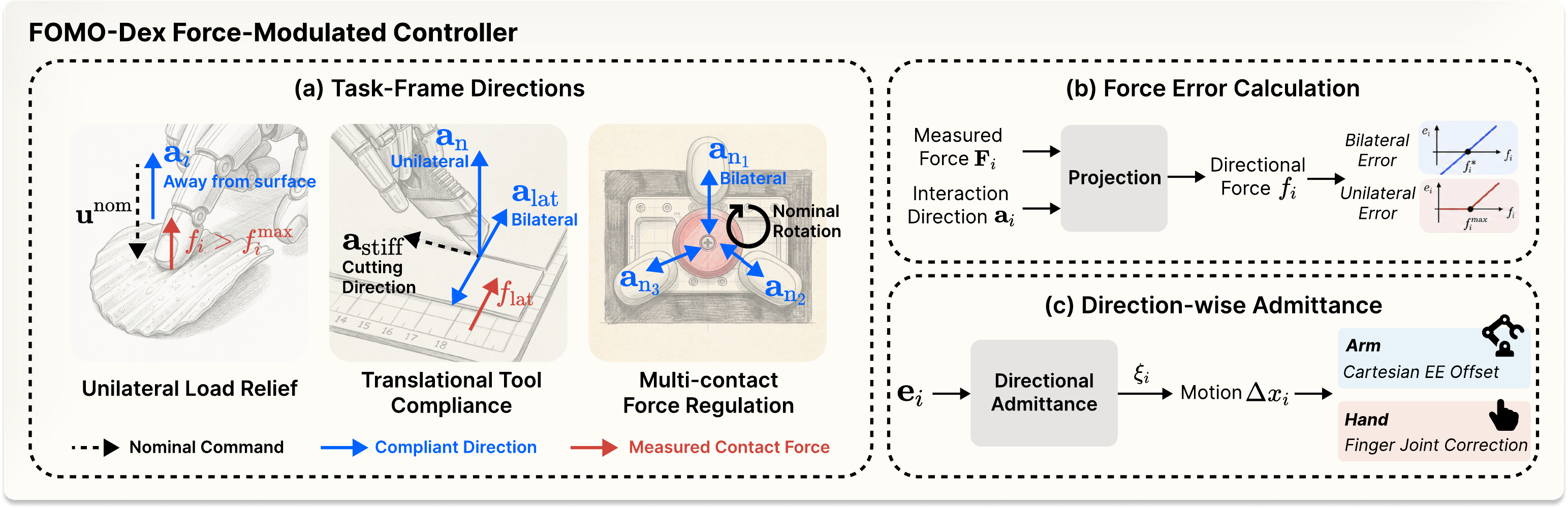}
\caption{Overview of the FOMO-Dex controller and its task-specific realizations.
Task geometry and contact structure define interaction directions, along which
force measurements are converted into unilateral or bilateral force errors.
Direction-wise admittance then produces arm or hand motion corrections,
enabling unilateral load relief, translational tool compliance, and multi-contact
force regulation within the same formulation.}
\label{fig:compliance-designs}
\end{figure}

To construct $\Delta\mathbf{u}_t$, we first choose interaction directions
from the task geometry and current contacts
(Section~\ref{section:task-frame-comp-rep}). We project force measurements
onto these directions and compare them with the corresponding targets or
limits. A direction-wise admittance law converts the resulting force errors
into motion corrections (Section~\ref{section:force-admittance}). We then
map these corrections to arm or finger commands
(Section~\ref{section:map-to-commands}). Figure~\ref{fig:compliance-designs}
illustrates this process and its task-specific realizations.

\subsubsection{Task-Frame Compliance Representation}
\label{section:task-frame-comp-rep}
At each time step $t$, we use the task geometry and contact structure to
define a set of $r$ unit interaction directions,
\begin{equation}
\mathcal A^{(t)} =
\{\mathbf a_1^{(t)},\ldots,\mathbf a_r^{(t)}\},
\qquad
\|\mathbf a_i^{(t)}\|_2=1.
\end{equation}

\noindent These directions identify the motion components relevant to the
interaction, following task-frame motion
decomposition~\cite{raibert1981hybrid,Bruyninckx1996SpecificationOF}. Each direction
$\mathbf a_i^{(t)}$ is associated with a contact-force measurement
$\mathbf F_i^{(t)}$ from the relevant sensor. The force component along
that direction is
\begin{equation}
f_i^{(t)}
=
\mathbf a_i^{(t)\mathsf T}\mathbf F_i^{(t)} .
\end{equation}
We denote the indices of the directions selected for force modulation by
$\mathcal C\subseteq\{1,\ldots,r\}$. Directions intended to remain stiff
are excluded from $\mathcal C$, so their nominal motion components receive
no compliance-induced correction.

\subsubsection{Force Error and Admittance Modulation}
\label{section:force-admittance}

For each selected direction $i\in\mathcal C$, we define a force error
relative to the desired contact condition. When the goal is to track a
target force $f_i^\star$, we use the signed error
\begin{equation}
e_i^{(t)}
=
f_i^{(t)} - f_i^\star .
\end{equation}
When the goal is only to limit excessive loading, we use a unilateral error
that is zero below the prescribed force limit $f_i^{\max}$,
\begin{equation}
e_i^{(t)}
=
\left[f_i^{(t)}-f_i^{\max}\right]_+ ,
\qquad
[x]_+ = \max(x,0).
\end{equation}

We convert the force error into a scalar motion correction $\xi_i$ through
a direction-wise admittance law~\cite{keemink2018admittance},
\begin{equation}
    m_i \ddot{\xi}_i
    +
    d_i \dot{\xi}_i
    +
    k_i \xi_i
    =
    e_i ,
    \qquad i\in\mathcal C,
    \label{eq:unified_admittance}
\end{equation}
where $m_i$, $d_i$, and $k_i$ are the virtual mass, damping, and stiffness
parameters for direction $\mathbf a_i$. This equation gives the
second-order mass--spring--damper response used for directional tool motion.
Setting $m_i=0$ gives a first-order response, while setting $m_i=k_i=0$
gives the integrating response used for other interactions.

Each admittance state $(\xi_i,\dot{\xi}_i)$ remains associated with the
same contact as its direction $\mathbf a_i^{(t)}$ changes. The scalar state
is retained, and the correction rotates with the direction. We reset the
state when a contact is created or lost, or when its direction enters or
leaves $\mathcal C$. This prevents corrections from carrying over between
contact modes.

We choose the sign of each interaction direction consistently with the
measured force, so that a positive force error drives a correction along
$\mathbf a_i$ that relieves excess contact load.

The resulting correction associated with interaction direction $i$ is
\begin{equation}
    \Delta\mathbf x_i
    =
    \xi_i\mathbf a_i .
    \label{eq:interaction_correction}
\end{equation}
Thus, task geometry determines \emph{where} force feedback acts, while the
admittance law determines \emph{how} force error modifies the nominal
motion.

\subsubsection{Mapping to Arm and Hand Commands}
\label{section:map-to-commands}

The task determines whether the arm or the fingers apply the
interaction-space corrections.

For arm-mediated compliance, let $\mathcal C_{\mathrm{arm}}$ contain the
indices of the directions assigned to the arm. We sum their corrections,
\begin{equation}
    \Delta\mathbf x_{\mathrm{arm}}
    =
    \sum_{i\in\mathcal C_{\mathrm{arm}}}
    \Delta\mathbf x_i ,
\end{equation}
and used to shift the Cartesian equilibrium pose supplied to the arm
impedance controller,
\begin{equation}
    \mathbf T_t^{\mathrm{cmd}}
    =
    \mathbf T_t^{\mathrm{nom}}
    \oplus
    \Delta\mathbf T_{\mathrm{arm}},
\end{equation}
where $\Delta\mathbf T_{\mathrm{arm}}$ is the pose correction corresponding
to $\Delta\mathbf x_{\mathrm{arm}}$.

For hand-mediated compliance, let $\mathcal C_j$ contain the indices of
the directions associated with fingertip $j$. Their corrections define the
desired fingertip displacement,
\begin{equation}
    \Delta\mathbf p_j
    =
    \sum_{i\in\mathcal C_j}
    \Delta\mathbf x_i ,
\end{equation}
We map these displacements to a joint correction $\Delta\mathbf q$ using
damped least-squares inverse kinematics~\cite{wampler1986manipulator}.
With fingertip Jacobian $\mathbf J_j$, candidate joint correction
$\mathbf z$, and regularization weight $\lambda$, we solve
\begin{equation}
    \Delta\mathbf q
    =
    \arg\min_{\mathbf z}
    \sum_j
    \left\|
        \mathbf J_j\mathbf z-\Delta\mathbf p_j
    \right\|_2^2
    +
    \lambda\|\mathbf z\|_2^2 ,
\end{equation}
and add $\Delta\mathbf q$ to the nominal finger-joint command as in
Eq.~\ref{eq:addition}.

Both mappings therefore use the same interaction-space correction, but
apply it through the command interface of the arm or hand.

\begin{table}[t]
\centering
\caption{Task-specific FOMO-Dex controller configurations. ``--'' denotes
not applicable. Contact on/off denotes the force thresholds used to activate and deactivate
contact-dependent modulation, respectively.}
\label{tab:controller_params}
\footnotesize
\begin{tabular}{@{}lccc@{}}
\toprule
 & Card Picking & Paper Cutting & Bolt--nut/Screwdriver \\
\midrule
Controller rate (Hz)        & 90 & 90 & 90 \\
Force target $f^\star$ (N)  & -- & -- & 0.5--2.4 \\
Force limit $f^{\max}$ (N)  & 3.0 & 3.5 & 3.0 \\
Contact on/off (N)          & 0.6/0.6 & 5.0/2.0 & 0.25/0.15 \\
Max correction              & $0.05$ rad & 0.06 m & 0.10 rad \\
\bottomrule
\end{tabular}

\end{table}

\subsubsection{Task-Specific Compliance Configuration}

We instantiate the controller for each task by specifying interaction
directions and their force targets or limits. Figure~\ref{fig:compliance-designs}(a)
shows the directions, and Table~\ref{tab:controller_params} lists the
controller settings. The three cases below differ in the contact condition
they regulate and whether the correction is applied through the arm or fingers.

\textbf{Unilateral load relief.}
During card picking, excessive fingertip force against the supporting
surface can disturb the pickup motion or damage the object. We choose an
interaction direction pointing away from the surface and use a unilateral
force error. Above the prescribed load limit, this error drives a retreat
correction through the finger joints. Below the limit, the force error is
zero. The correction thus responds to excessive surface loading without
requiring the fingertip to maintain a fixed contact force.

\textbf{Translational tool compliance.}
When cutting paper with a utility knife, a mismatch between the commanded
motion and blade orientation can generate lateral forces that deflect the
blade or tear the paper. We exclude the cutting direction from force
modulation and apply compliance in the lateral and surface-normal
directions. Bilateral feedback lets the arm respond to lateral forces from
either side. Along the surface normal, a unilateral force error drives
retreat when the pressing force exceeds its limit. This leaves the commanded
cutting motion unchanged while reducing lateral and excessive normal loads.

\textbf{Multi-contact force regulation.}
During rotational manipulation, the fingers must maintain contact as they
move around the rotation axis. Small errors in the nominal motion can cause
contact loss or excessive compression. We use the local fingertip contact
normals $\mathbf n_i$ as interaction directions and assign a compressive
force target $f_i^\star$ to each contact. These targets are derived from a
shared grip level $g$, which we slowly adapt based on contact loss, friction
utilization, and excessive loading. We distribute this grip level across
active contacts according to the current grasp geometry, favoring equal
loading while minimizing the net wrench induced by the contact-normal
forces. Bilateral admittance increases compression when the measured force
is below its target and reduces compression when the force is too large.
The commanded rotational motion is left unchanged while the controller
adjusts the fingertip loads.

\subsection{Visuomotor Policy Training}
We train a visuomotor policy by behavior cloning to evaluate whether the
collected demonstrations support autonomous manipulation. The policy takes
RGB images from the head ZED camera together with robot proprioception and
force measurements. A frozen pretrained DINOv2 encoder~\cite{oquab2023dinov2}
extracts image features, and a diffusion policy~\cite{chi2023diffusion}
predicts joint-space actions.

The training targets are the force-modulated joint commands sent to the
robot, rather than the nominal teleoperation commands. The policy therefore
learns from both the operator's motion and the force-dependent corrections
in the demonstrations. During training, we randomly mask the proprioceptive
inputs to encourage the policy to use visual and force observations.

\section{Experiments}
\label{sec:experiment}
We evaluate FOMO-Dex in teleoperation and autonomous policy execution through
three questions:

1) \textit{How does force modulation affect contact forces and task progress during teleoperation?}

2) \textit{Across different users, does our controller increase successful-demonstration throughput relative to stiff control?}

3) \textit{How do demonstration collection and action targets affect
    autonomous policy performance?}

\subsection{Experimental Setup}

\subsubsection{Baselines}
We compare our force-modulated controller described in
Section~\ref{sec:method} with the following baselines. The \textbf{Stiff} controller directly executes the
operator command with hand position control and arm impedance control, i.e., $\Delta\mathbf u_t=0$ in Eq.~\ref{eq:addition}. The
\textbf{Naive Compliant} controller uses the same admittance parameters as
FOMO-Dex wherever applicable, but applies them isotropically rather than
selectively along task- and contact-relevant directions. All controllers share the same teleoperation interface,
command rate, sensing, low-level control, and safety limits. 

\begin{table}[t]
\centering
\caption{
\textbf{Evaluation of our force-modulated controller across four benchmark
tasks compared with stiff and naive-compliance baselines over 50 trials.} 
Exc. Frames denotes the fraction
of frames in which the measured contact force exceeds the task-specific
safe threshold, and Force Exceedance is defined in
Eq.~\ref{eq:force_exceedance}. $\uparrow$/$\downarrow$ indicate
higher/lower is better; bold and underline denote the best and second-best
results, respectively. 
}
\label{tab:controller_comparison}

\footnotesize
\setlength{\tabcolsep}{5pt}
\renewcommand{\arraystretch}{1.1}

\begin{tabular}{@{}llccc@{}}
\toprule
\textbf{Task} & \textbf{Metric}
& \textbf{Stiff}
& \textbf{Naive Comp.}
& \textbf{Ours} \\
\midrule

\multirow{4}{*}{Card Picking}
& Success (\%) $\uparrow$
& \underline{90.9} & 37.5 & \textbf{93.8} \\
& Duration (s) $\downarrow$
& 22.1{\scriptsize $\pm 4.2$} & \underline{17.2}{\scriptsize $\pm 2.4$} & \textbf{16.9}{\scriptsize $\pm 4.2$} \\
& Exc. Frames (\%) $\downarrow$
& 13.0{\scriptsize $\pm 5.4$} & \underline{7.4}{\scriptsize $\pm 7.9$} & \textbf{5.9}{\scriptsize $\pm 2.8$} \\
& Force Exceedance (N) $\downarrow$
& 0.32{\scriptsize $\pm 0.29$} & \underline{0.10}{\scriptsize $\pm 0.13$} & \textbf{0.05}{\scriptsize $\pm 0.03$} \\

\midrule

\multirow{3}{*}{Paper Cutting}
& Cut Rate (cm/s) $\uparrow$
& 2.47{\scriptsize $\pm 0.15$} & \textbf{3.16}{\scriptsize $\pm 1.12$} & \underline{2.63}{\scriptsize $\pm 0.41$} \\
& Lat. Force (N) $\downarrow$
& 2.00{\scriptsize $\pm 0.41$} & \underline{1.61}{\scriptsize $\pm 0.43$} & \textbf{1.41}{\scriptsize $\pm 0.14$} \\
& Norm. Force (N) $\downarrow$
& \underline{4.39}{\scriptsize $\pm 0.94$} & 5.28{\scriptsize $\pm 0.77$} & \textbf{3.62}{\scriptsize $\pm 0.18$} \\

\midrule

\multirow{3}{*}{Bolt-Nut}
& Turns/ep $\uparrow$
& \textbf{23.7}{\scriptsize $\pm 1.2$} & 17.0{\scriptsize $\pm 2.1$} & \underline{20.3}{\scriptsize $\pm 1.2$} \\
& Exc. Frames (\%) $\downarrow$
& 27.4{\scriptsize $\pm 4.1$} & \textbf{8.0}{\scriptsize $\pm 2.0$} & \underline{23.3}{\scriptsize $\pm 4.3$} \\
& Force Exceedance (N) $\downarrow$
& 0.64{\scriptsize $\pm 0.18$} & \textbf{0.09}{\scriptsize $\pm 0.03$} & \underline{0.57}{\scriptsize $\pm 0.21$} \\

\midrule

\multirow{4}{*}{Screwdriving}
& Turns/ep $\uparrow$
& 4.2{\scriptsize $\pm 2.2$} & \underline{14.6}{\scriptsize $\pm 6.9$} & \textbf{34.7}{\scriptsize $\pm 5.7$} \\
& Slip Events/ep $\downarrow$
& 1.00{\scriptsize $\pm 0.00$} & \underline{0.75}{\scriptsize $\pm 0.43$} & \textbf{0.28}{\scriptsize $\pm 0.46$} \\
& Exc. Frames (\%) $\downarrow$
& 0.4{\scriptsize $\pm 0.9$} & \textbf{0.2}{\scriptsize $\pm 0.2$} & \underline{0.3}{\scriptsize $\pm 1.2$} \\
& Force Exceedance (N) $\downarrow$
& 0.186{\scriptsize $\pm 0.073$} & \textbf{0.001}{\scriptsize $\pm 0.001$} & \underline{0.002}{\scriptsize $\pm 0.009$} \\

\bottomrule
\end{tabular}
\end{table}

\subsubsection{Benchmark Problems}
We evaluate our method on four contact-rich dexterous manipulation tasks that require different forms of force-aware compliance. 
We report task-progress metrics and interaction-based metrics. 
Let $f_t$ denote the
relevant measured contact force and $f^{\mathrm{safe}}$ its task-dependent
threshold. We report the fraction of frames exceeding this threshold and the
mean force exceedance
\begin{equation}
    E_f = \frac{1}{T}\int_0^T
    \max\!\left(0, |f_t|-f^{\mathrm{safe}}\right)\,dt,
    \label{eq:force_exceedance}
\end{equation}
where $T$ is the episode duration. This metric captures both the magnitude
and duration of force above the threshold.

\paragraph{Card Picking}
The robot picks a poker card from a flat surface while avoiding excessive
fingertip loading against the support. We measure pickup success, episode
duration, the fraction of frames with excessive contact force, and force
exceedance using Eq.~\ref{eq:force_exceedance}.

\paragraph{Paper Cutting}
The robot uses a utility knife to follow a marked path through a sheet of
paper while maintaining appropriate contact with the surface. We measure
cut rate and the lateral and normal interaction forces applied to the tool.

\paragraph{Bolt--Nut Screwing}
The robot rotates a nut along a threaded bolt while maintaining fingertip
contact. We measure the number of completed turns per episode, the fraction
of frames with excessive contact force, and force exceedance using
Eq.~\ref{eq:force_exceedance}.

\paragraph{Screwdriving}
The robot turns a screw with a screwdriver while maintaining tip engagement
and axial contact. We measure the number of completed turns per episode,
screwdriver slip events, the fraction of frames with excessive contact
force, and force exceedance using Eq.~\ref{eq:force_exceedance}.

\subsubsection{Trial protocol}
For each task, the object and robot are reset to a common initial
configuration. We record operator commands, post-compliance commands, robot
state, wrist wrench and fingertip forces at 30 Hz. The thresholds used to compute excessive
force are fixed during data collection.

\subsection{Controller Analysis on Teleoperation Data}
\label{sec:controller_results}

\begin{figure}[t]
    \centering
    \includegraphics[width=0.75\linewidth]{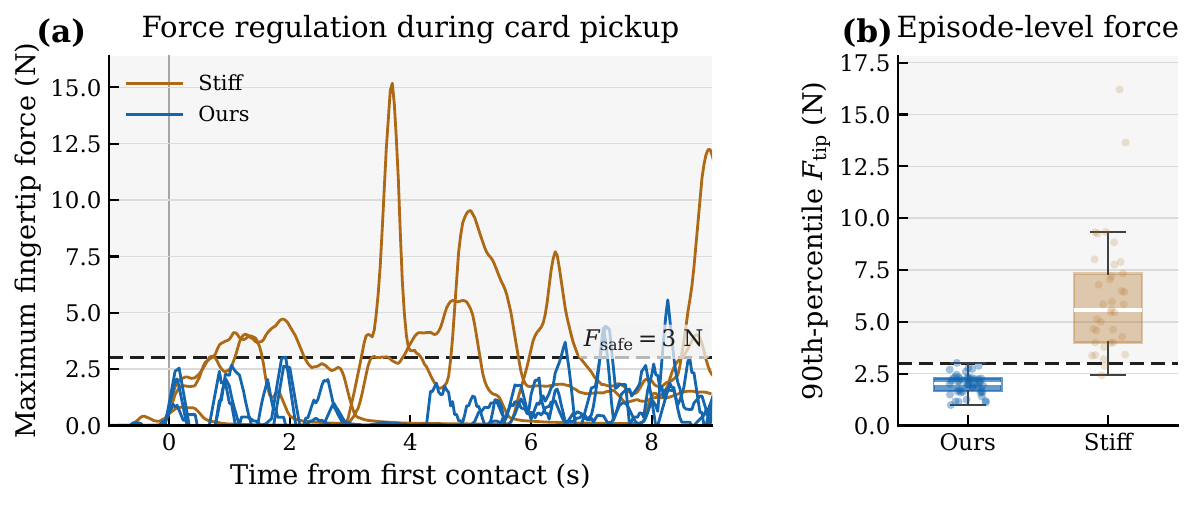}
    
    \caption{Fingertip forces during card picking with stiff control and FOMO-Dex.
(a) Maximum force across fingertips over five representative episodes per
controller, aligned at first contact. (b) Distribution of the per-episode
90th-percentile fingertip force. Dashed lines mark the 3~N force threshold.}
    
    \label{fig:card-pickup-force}
\end{figure}

\begin{figure}[!ht]
    \centering
    \includegraphics[width=0.75\linewidth]{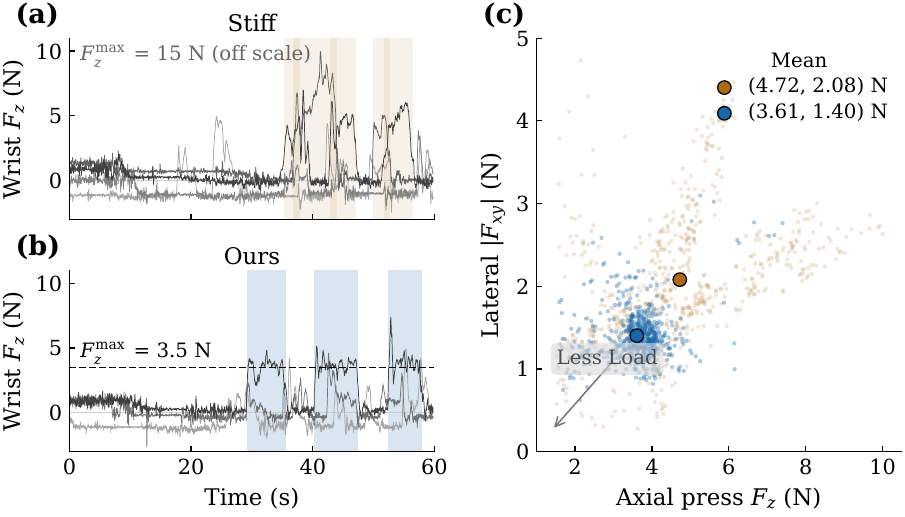}
    \caption{Interaction forces during paper cutting.
(a,b) Wrist axial force over five representative episodes under stiff control
and FOMO-Dex, respectively. (c) Axial versus lateral force samples for stiff
control (orange) and FOMO-Dex (blue). Large markers indicate the means.}
    \label{fig:cutting-force}
\end{figure}

\begin{figure}[!ht]
    \centering
    \includegraphics[width=0.49\linewidth,trim=0 284bp 0 0,clip]{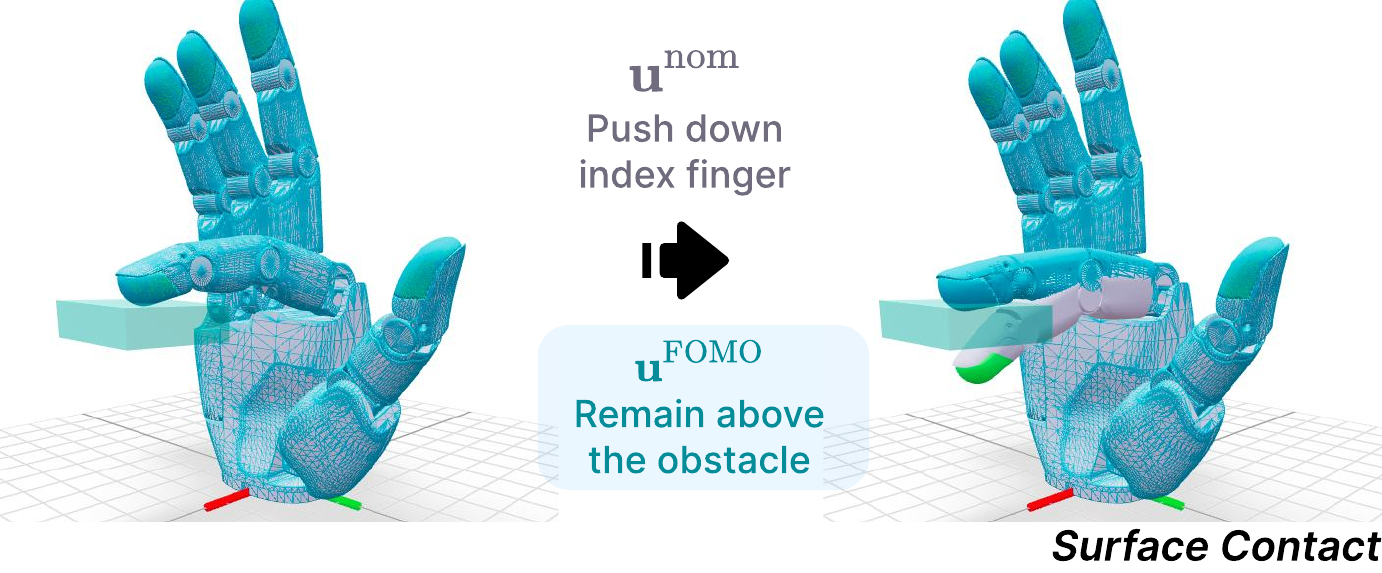}%
    \hfill
    \includegraphics[width=0.49\linewidth,trim=0 0 0 300bp,clip]{figures/sim-vis.pdf}
    
    \caption{Hand configurations before and after force modulation. The gray mesh
shows the nominal configuration commanded by the teleoperator, while the blue
mesh shows the force-modulated command.
For surface contact (left), the contacting fingertip retreats from the
supporting surface. For in-hand tripod contact (right), force regulation adjusts the
index--thumb contacts while preserving the remaining grasp configuration.}
    \label{fig:sim-vis}
    
\end{figure}

\begin{figure}[t]
    \centering
    \includegraphics[width=\linewidth]{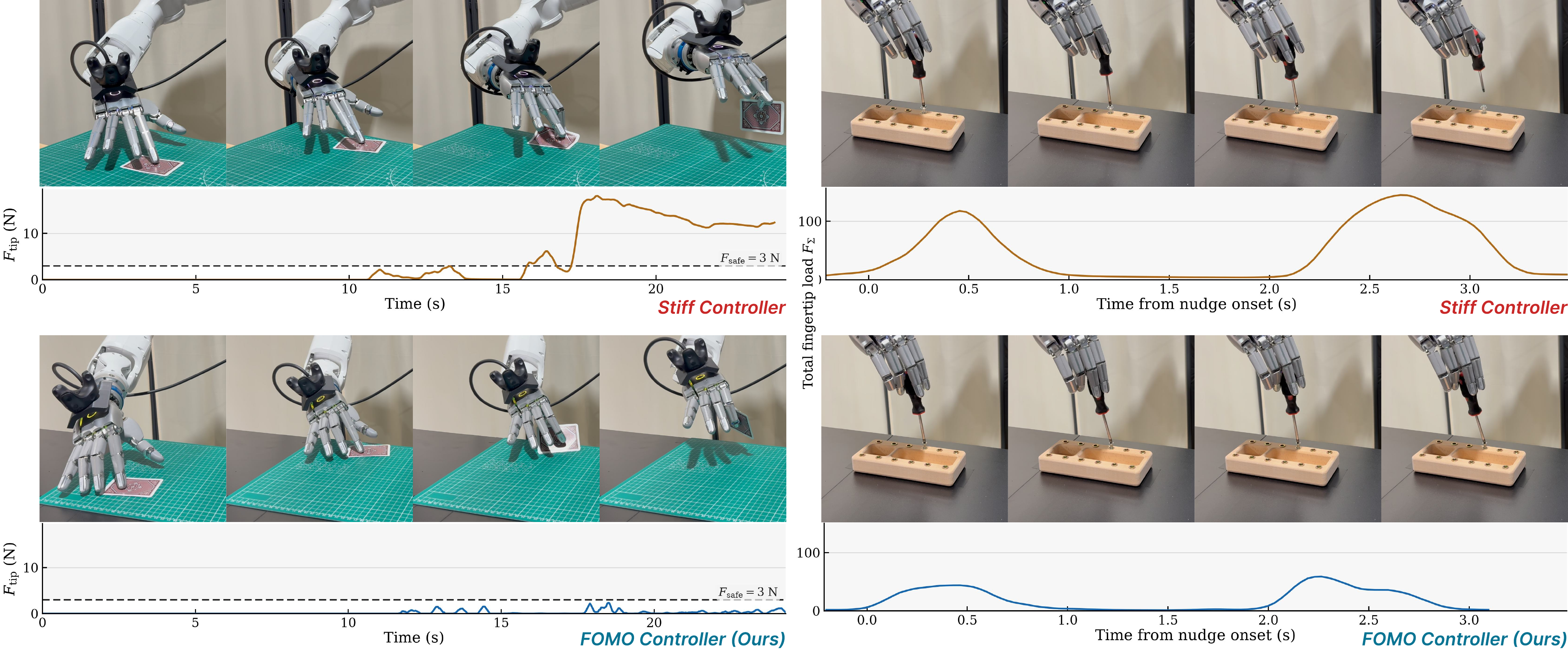}
    \caption{Real-world teleoperation with stiff control (top) and FOMO-Dex
(bottom). In card picking (left), FOMO-Dex limits excessive fingertip
loading against the supporting surface. In screwdriving (right), it reduces
large variations in total fingertip load and maintains more stable tool
engagement. Representative execution frames are shown with the corresponding
force traces.}
    \label{fig:teleop-rollout}
    
\end{figure}

We first evaluate how force modulation affects the execution of nominal
teleoperation commands across four contact-rich tasks.
Table~\ref{tab:controller_comparison} compares our task-frame
force-modulated controller against baseline controllers. Stiff control tracks
the operator command without explicit force-dependent corrections. This leads to excessive
interaction forces in tasks such as card picking and paper cutting, and can
also destabilize contact in screwdriving. Naive compliance alleviates some of
these forces, but indiscriminately relaxing the commanded motion can interfere
with task progress. This is particularly evident in card picking, where reduced
contact force does not translate into reliable task completion.

The balance between force reduction and task progress differs across tasks.
For \textit{card picking}, the controller can yield to the supporting surface while
preserving the motion needed to retrieve the card. Figure~\ref{fig:card-pickup-force} overlays fingertip forces across five card-pickup episodes, showing that our controller consistently limits excessive contact forces, while stiff control exhibits frequent high-force interactions.
During \textit{paper cutting}, FOMO-Dex reduces lateral and normal loads
relative to stiff control while maintaining a comparable cut rate.
Figure~\ref{fig:cutting-force} shows these force reductions across five episodes.
In \textit{bolt-nut manipulation}, stiff control favors rotational progress at the
cost of larger contact forces, while naive compliance reduces contact forces
but more strongly suppresses the commanded rotation. FOMO-Dex lies between
these baselines in both completed turns and force exceedance.
In \textit{screwdriving}, force
modulation substantially improves rotational progress and maintains tool
engagement. Naive compliance achieves slightly lower force exceedance than
FOMO-Dex, but completes fewer turns and has more tool slips. Lower force
exceedance alone therefore does not imply better manipulation performance.

To further illustrate how force modulation changes the commanded hand
configuration, Fig.~\ref{fig:sim-vis} visualizes representative
contact-dependent corrections. The gray mesh denotes the nominal hand
configuration, while the blue mesh shows the force-modulated configuration.
For surface contact, the contacting fingertip retreats from the obstacle while
the remaining fingers largely preserve the nominal motion. Under in-hand
multi-contact interaction, the controller instead adjusts the active contacts
while maintaining the overall grasp. These examples highlight that FOMO-Dex
does not uniformly soften the hand, but selectively modifies motion components
associated with the current contact constraints.

Figure~\ref{fig:teleop-rollout} shows how these contact-dependent corrections
translate to real-world teleoperation behavior. In card picking, stiff control
produces a large fingertip-force spike as the hand interacts with the supporting
surface, whereas FOMO-Dex relieves the contact load while allowing the pickup
motion to continue. In screwdriving, stiff control exhibits large variations in
total fingertip load, while FOMO-Dex attenuates these variations and maintains
more stable tool engagement.

\subsection{User Study}
\label{sec:user_study}

\begin{table}[t]
\centering
\caption{
\textbf{User-study comparison of demonstration collection with the stiff controller and FOMO-Dex, zero-shot and after 10 min of practice.}
Succ. Data is successful demonstration time per hour of teleoperation.
Mean Ep. Len. is episode duration; shorter is preferred for card picking, while longer is preferred for screwdriving, where episodes are capped at 60 s and longer durations indicate more stable sustained manipulation of the long screw.
}
\label{tab:user_study}
\footnotesize
\setlength{\tabcolsep}{5pt}
\renewcommand{\arraystretch}{1.1}

\begin{tabular}{@{}lcccc@{}}
\toprule
&
\multicolumn{2}{c}{\textbf{Stiff}} &
\multicolumn{2}{c}{\textbf{Ours}} \\
\cmidrule(lr){2-3}
\cmidrule(lr){4-5}

\textbf{Experience} &
\textbf{Succ. Data} &
\textbf{Mean Ep. Len.} &
\textbf{Succ. Data} &
\textbf{Mean Ep. Len.} \\
&
\textbf{(min/h) $\uparrow$} &
\textbf{(s)} &
\textbf{(min/h) $\uparrow$} &
\textbf{(s)} \\
\midrule

\multicolumn{5}{@{}l}{\textbf{Card Picking} \quad
\textit{Mean Ep. Len. $\downarrow$}} \\
Zero-shot
& 11.9{\scriptsize $\pm 5.2$} & 32.7{\scriptsize $\pm 10.5$}
& 20.6{\scriptsize $\pm 4.0$} & 29.1{\scriptsize $\pm 0.5$} \\
+10 min
& 24.3{\scriptsize $\pm 9.4$} & 27.9{\scriptsize $\pm 1.2$}
& \textbf{29.3}{\scriptsize $\pm 5.6$} & \textbf{24.9}{\scriptsize $\pm 10.2$} \\

\midrule

\multicolumn{5}{@{}l}{\textbf{Bolt-Nut} \quad
\textit{Mean Ep. Len. = 60}} \\
Zero-shot
& 41.0{\scriptsize $\pm 1.1$} & 60.0{\scriptsize $\pm 0.0$}
& 40.7{\scriptsize $\pm 1.0$} & 60.0{\scriptsize $\pm 0.0$} \\
+10 min
& 30.3{\scriptsize $\pm 12.8$} & 60.0{\scriptsize $\pm 0.0$}
& \textbf{50.2}{\scriptsize $\pm 7.0$} & 60.0{\scriptsize $\pm 0.0$} \\

\midrule

\multicolumn{5}{@{}l}{\textbf{Screwdriving} \quad
\textit{Mean Ep. Len. $\uparrow$}} \\
Zero-shot
& 11.8{\scriptsize $\pm 6.4$} & \textbf{11.2}{\scriptsize $\pm 5.2$}
& 17.9{\scriptsize $\pm 15.2$} & 10.9{\scriptsize $\pm 6.8$} \\
+10 min
& 12.4{\scriptsize $\pm 4.7$} & 6.4{\scriptsize $\pm 2.0$}
& \textbf{30.7}{\scriptsize $\pm 10.9$} & 10.4{\scriptsize $\pm 2.3$} \\

\bottomrule
\end{tabular}

\end{table}

We conduct a user study with four participants to evaluate how force
modulation affects teleoperated demonstration collection. Each participant
performs three tasks---card picking, bolt--nut manipulation, and
screwdriving---using both our controller and the stiff controller. Task
order and controller order are randomized across participants.

For each task--controller condition, participants first perform 10 trials
without prior practice (zero-shot). They are then given 10 minutes to
practice the task using both controllers, after which they perform another
10 evaluation trials with each controller. Participants receive the same
task instructions and success criteria across conditions.

We measure the amount of successful demonstration data collected per hour
of teleoperation (Succ. Data) and the mean episode length. Succ. Data is
computed as the total duration of successful demonstrations divided by the
total teleoperation time, and scaled to minutes of successful data per hour
of human teleoperation.

Table~\ref{tab:user_study} shows that FOMO-Dex increases successful-data
throughput across all three tasks after practice. For card picking, our controller
increases successful-data throughput both zero-shot and after practice.
For bolt--nut manipulation, the two controllers perform similarly
zero-shot, while after practice participants collect substantially more
successful data with FOMO-Dex. In screwdriving, FOMO-Dex improves throughput
both before and after practice. After practice, participants also sustain
longer episodes with FOMO-Dex than with stiff control.

\subsection{Behavior-Cloning Results}
\label{sec:bc_results}

\begin{table}[t]
\centering
\caption{
\textbf{Evaluation of behavior-cloning policies trained on 50 demonstrations per task--controller setting.}
(a) Card-picking ablation comparing stiff, raw-command training with FOMO-Dex inference
(Comp. Raw), and post-modulation-action training (Comp. Applied) over 18 evaluation trials.
(b) Cross-task comparison of policies trained from stiff and FOMO-Dex
demonstrations, evaluated by turns per episode, over 10 evaluation trials.
}
\label{tab:bc_results}
\footnotesize

\begin{minipage}[t]{0.52\columnwidth}
\centering
\textbf{(a) Card Picking Ablation}\\[1pt]

\setlength{\tabcolsep}{5pt}
\begin{tabular}{@{}lccc@{}}
\toprule
\textbf{Success (\%)} & \textbf{Reach} & \textbf{Retr.} & \textbf{Lift} \\
\midrule
Stiff              & 100 & 100 & 56 \\
Comp. Raw        & 100 & 89  & 0 \\
Comp. Applied (Ours)    & 100 & 89  & \textbf{67} \\
\bottomrule
\end{tabular}
\end{minipage}
\hfill
\begin{minipage}[t]{0.43\columnwidth}
\centering
\textbf{(b) Cross-task Comparison}\\[2pt]

\setlength{\tabcolsep}{5pt}
\begin{tabular}{@{}lcc@{}}
\toprule
\textbf{Turns/ep.} & \textbf{Stiff} & \textbf{Ours} \\
\midrule
Bolt-Nut
& 24.5{\scriptsize $\pm 4.2$}
& \textbf{60.8}{\scriptsize $\pm 4.5$} \\
Screwdriving
& 0.6{\scriptsize $\pm 0.8$}
& \textbf{6.3}{\scriptsize $\pm 2.2$} \\
\bottomrule
\end{tabular}
\end{minipage}

\end{table}

We further investigate how force-modulated demonstrations should be used for policy learning. We consider two ways of incorporating our controller into behavior
cloning. \emph{Comp. Raw} trains the policy to predict the raw teleoperation
commands and applies our force-modulation controller at inference time.
\emph{Comp. Applied} instead uses force-modulated actions as training targets
and applies the stiff controller at inference time.

We first compare these strategies on card picking.
As shown in Table~\ref{tab:bc_results}(a), directly imitating the applied actions produces the highest lift success. The policy trained on raw commands completes no successful lifts, even with force modulation at deployment.
Based on this result, we use post-modulation actions as the learning target in the remaining tasks.

We next evaluate whether demonstrations collected with force modulation improve policy learning on long-horizon rotational tasks.
We compare policies trained on demonstrations collected with stiff teleoperation against those collected with FOMO-Dex, using the applied actions as targets for the latter. As shown in Table~\ref{tab:bc_results}(b), policies trained on FOMO-Dex demonstrations achieve substantially more rotation in both bolt--nut manipulation and screwdriving.

\section{Conclusions}
We present FOMO-Dex, a unified framework for force-modulated dexterous manipulation that treats teleoperation or policy output as a nominal action and selectively adjusts contact-relevant components using force feedback in task- and contact-relevant coordinates. Across four contact-rich tasks, FOMO-Dex improves contact behavior while preserving task progress, increases successful demonstration throughput in a user study, and enables behavior-cloning policies to benefit from force-modulated demonstrations. These results suggest that explicit force modulation provides a practical interface between motion commands and the contact responses required for dexterous manipulation. Future work will infer interaction directions and force targets online, incorporate uncertainty-aware contact estimation, and evaluate the framework across broader tasks and objects.

\bibliographystyle{plainnat}
\bibliography{root}

@article{abu2018force,
  title={Force-based variable impedance learning for robotic manipulation},
  author={Abu-Dakka, Fares J and Rozo, Leonel and Caldwell, Darwin G},
  journal={Robotics and Autonomous Systems},
  year={2018},
}

@inproceedings{zeng2021learning,
  title={Learning compliant grasping and manipulation by teleoperation with adaptive force control},
  author={Zeng, Chao and Li, Shuang and Jiang, Yiming and Li, Qiang and Chen, Zhaopeng and Yang, Chenguang and Zhang, Jianwei},
  booktitle={IROS},
  year={2021},
}

@article{abu2020variable,
  title={Variable impedance control and learning---a review},
  author={Abu-Dakka, Fares J and Saveriano, Matteo},
  journal={Frontiers in Robotics and AI},
  year={2020},
}

@inproceedings{zhang2022learning,
  title={Learning and generalizing variable impedance manipulation skills from human demonstrations},
  author={Zhang, Yan and Zhao, Fei and Liao, Zhiwei},
  booktitle={AIM},
  year={2022},
}

@article{zhou2025variable,
  title={Variable impedance control on contact-rich manipulation of a collaborative industrial mobile manipulator: An imitation learning approach},
  author={Zhou, Zhengxue and Yang, Xingyu and Zhang, Xuping},
  journal={Robotics and Computer-Integrated Manufacturing},
  year={2025},
}

@inproceedings{shukla2025force,
  title={Force-conditioned diffusion policies for compliant sheet separation tasks in bimanual robotic cells},
  author={Shukla, Rishabh and Talan, Raj and Moode, Samrudh and Dhanaraj, Neel and Kang, Jeon Ho and Gupta, Satyandra K},
  booktitle={ICRA},
  year={2025},
}

@article{kang2025robotic,
  title={Robotic compliant object prying using diffusion policy guided by vision and force observations},
  author={Kang, Jeon Ho and Joshi, Sagar and Huang, Ruopeng and Gupta, Satyandra K},
  journal={RA-L},
  year={2025},
}

@inproceedings{hou2025adaptive,
  title={Adaptive compliance policy: Learning approximate compliance for diffusion guided control},
  author={Hou, Yifan and Liu, Zeyi and Chi, Cheng and Cousineau, Eric and Kuppuswamy, Naveen and Feng, Siyuan and Burchfiel, Benjamin and Song, Shuran},
  booktitle={ICRA},
  year={2025},
}

@inproceedings{shi2026minimalist,
  title={Minimalist compliance control},
  author={Shi, Haochen and Hu, Songbo and Hou, Yifan and Wang, Weizhuo and Liu, Karen and Song, Shuran},
  booktitle={RSS},
  year={2026}
}

@inproceedings{xu2026compliant,
  title={Compliant residual {DAgger}: Improving real-world contact-rich manipulation with human corrections},
  author={Xu, Xiaomeng and Hou, Yifan and Liu, Zeyi and Song, Shuran},
  booktitle={NeurIPS},
  year={2025}
}

@inproceedings{choi2026wild,
  title={In-the-wild compliant manipulation with {UMI-FT}},
  author={Choi, Hojung and Hou, Yifan and Pan, Chuer and Hong, Seongheon and Patel, Austin and Xu, Xiaomeng and Cutkosky, Mark R and Song, Shuran},
  booktitle={ICRA},
  year={2026}
}

@article{khalil2026vidp,
  title={{VIDP}: Variable impedance diffusion policy for compliant robot manipulation from diverse demonstrations},
  author={Khalil, Hisham and Fernandes, Neil and Kwok, Thomas M and Lin, Hsiu-Chin and Hu, Yue},
  journal={arXiv:2608.06210},
  year={2026}
}

@article{liu2026dexteleop,
  title={{DexTeleop-0}: Force-aware bimanual dexterous teleoperation with ego-centric perception towards shared autonomy},
  author={Liu, Haichao and Jiang, Yuyao and Park, Hyunsun and Xue, Yuanjiang and Wang, Ziwei},
  journal={arXiv:2606.23431},
  year={2026}
}

@article{Bruyninckx1996SpecificationOF,
  title={Specification of force-controlled actions in the "task frame formalism"-a synthesis},
  author={Bruyninckx, Herman and De Schutter, Joris},
  journal={IEEE Transactions on Robotics and Automation},
  year={1996},
}

@inproceedings{hogan1984impedance,
  title={Impedance control: An approach to manipulation},
  author={Hogan, Neville},
  booktitle={ACC},
  year={1984},
}

@inproceedings{ding2025bunny,
  title={{Bunny-VisionPro}: Real-time bimanual dexterous teleoperation for imitation learning},
  author={Ding, Runyu and Qin, Yuzhe and Zhu, Jiyue and Jia, Chengzhe and Yang, Shiqi and Yang, Ruihan and Qi, Xiaojuan and Wang, Xiaolong},
  booktitle={IROS},
  year={2025},
}

@inproceedings{shaw2024bimanual,
  title={Bimanual dexterity for complex tasks},
  author={Shaw, Kenneth and Li, Yulong and Yang, Jiahui and Srirama, Mohan Kumar and Liu, Ray and Xiong, Haoyu and Mendonca, Russell and Pathak, Deepak},
  booktitle={CoRL},
  year={2024}
}

@article{qin2022one,
  title={From one hand to multiple hands: Imitation learning for dexterous manipulation from single-camera teleoperation},
  author={Qin, Yuzhe and Su, Hao and Wang, Xiaolong},
  journal={RA-L},
  year={2022},
}

@inproceedings{handa2020dexpilot,
  title={{DexPilot}: Vision-based teleoperation of dexterous robotic hand-arm system},
  author={Handa, Ankur and Van Wyk, Karl and Yang, Wei and Liang, Jacky and Chao, Yu-Wei and Wan, Qian and Birchfield, Stan and Ratliff, Nathan and Fox, Dieter},
  booktitle={ICRA},
  year={2020},
}

@inproceedings{qin2023anyteleop,
  title     = {{AnyTeleop}: A general vision-based dexterous robot arm-hand teleoperation system},
  author    = {Qin, Yuzhe and Yang, Wei and Huang, Binghao and Van Wyk, Karl and Su, Hao and Wang, Xiaolong and Chao, Yu-Wei and Fox, Dieter},
  booktitle = {RSS},
  year      = {2023}
}

@inproceedings{yang2024ace,
  title={{ACE}: A cross-platform and visual-exoskeletons system for low-cost dexterous teleoperation},
  author={Yang, Shiqi and Liu, Minghuan and Qin, Yuzhe and Ding, Runyu and Li, Jialong and Cheng, Xuxin and Yang, Ruihan and Yi, Sha and Wang, Xiaolong},
  booktitle={CoRL},
  year={2024}
}

@INPROCEEDINGS{ZhangH-RSS-25, 
    AUTHOR    = {Han Zhang AND Songbo Hu AND Zhecheng Yuan AND Huazhe Xu}, 
    TITLE     = {{DOGlove}: Dexterous manipulation with a low-cost open-source haptic force feedback glove},
    BOOKTITLE = {RSS},
    YEAR      = {2025}, 
}

@article{oquab2023dinov2,
  title={{DINOv2}: Learning robust visual features without supervision},
  author={Oquab, Maxime and Darcet, Timoth{\'e}e and Moutakanni, Th{\'e}o and Vo, Huy and Szafraniec, Marc and Khalidov, Vasil and Fernandez, Pierre and Haziza, Daniel and Massa, Francisco and El-Nouby, Alaaeldin and others},
  journal={TMLR},
  year={2024}
}

@inproceedings{chi2023diffusion,
	title={Diffusion Policy: Visuomotor Policy Learning via Action Diffusion},
	author={Chi, Cheng and Feng, Siyuan and Du, Yilun and Xu, Zhenjia and Cousineau, Eric and Burchfiel, Benjamin and Song, Shuran},
	booktitle={RSS},
	year={2023}
}

@article{raibert1981hybrid,
  title={Hybrid position/force control of manipulators},
  author={Raibert, Marc H and Craig, John J},
  journal={Journal of Dynamic Systems, Measurement, and Control},
  year={1981}
}

@article{keemink2018admittance,
  title={Admittance control for physical human--robot interaction},
  author={Keemink, Arvid QL and Van der Kooij, Herman and Stienen, Arno HA},
  journal={IJRR},
  year={2018},
}

@article{wampler1986manipulator,
  title={Manipulator inverse kinematic solutions based on vector formulations and damped least-squares methods},
  author={Wampler, Charles W},
  journal={IEEE Transactions on Systems, Man, and Cybernetics},
  year={1986},
}

@inproceedings{li2014object,
  title={Learning object-level impedance control for robust grasping and dexterous manipulation},
  author={Li, Miao and Yin, Hang and Tahara, Kenji and Billard, Aude},
  booktitle={ICRA},
  year={2014}
}

@article{pfanne2020object,
  title={Object-level impedance control for dexterous in-hand manipulation},
  author={Pfanne, Martin and Chalon, Maxime and Stulp, Freek and Ritter, Helge and Albu-Sch{\"a}ffer, Alin},
  journal={RA-L},
  year={2020}
}

@inproceedings{hou2021efficient,
  title={An efficient closed-form method for optimal hybrid force-velocity control},
  author={Hou, Yifan and Mason, Matthew T},
  booktitle={ICRA},
  year={2021}
}

@inproceedings{xu2025dexumi,
  title={{DexUMI}: Using human hand as the universal manipulation interface for dexterous manipulation},
  author={Xu, Mengda and Zhang, Han and Hou, Yifan and Xu, Zhenjia and Fan, Linxi and Veloso, Manuela and Song, Shuran},
  booktitle={CoRL},
  year={2025}
}

@article{chen2025dexforce,
  title={{DexForce}: Extracting force-informed actions from kinesthetic demonstrations for dexterous manipulation},
  author={Chen, Claire and Yu, Zhongchun and Choi, Hojung and Cutkosky, Mark and Bohg, Jeannette},
  journal={RA-L},
  year={2025}
}

@inproceedings{xu2026contact,
  title={Contact-grounded policy: Dexterous visuotactile policy with generative contact grounding},
  author={Xu, Zhengtong and Wang, Yeping and Abbatematteo, Ben and Preechayasomboon, Jom and Chan, Sonny and Colonnese, Nicholas and Memar, Amirhossein H},
  booktitle={RSS},
  year={2026}
}

@inproceedings{fang2026forcepolicy,
  title={Force policy: Learning hybrid force-position control policy under interaction frame for contact-rich manipulation},
  author={Fang, Hongjie and Tang, Shirun and Mei, Mingyu and Qin, Haoxiang and He, Zihao and Chen, Jingjing and Feng, Ying and Wang, Chenxi and Liu, Wanxi and He, Zaixing and Lu, Cewu and Wang, Shiquan},
  booktitle={RSS},
  year={2026}
}

@inproceedings{hsieh2026dexscrew,
  title={Learning dexterous manipulation skills from imperfect simulations},
  author={Hsieh, Elvis and Hsieh, Wen-Han and Wang, Yen-Jen and Lin, Toru and Malik, Jitendra and Sreenath, Koushil and Qi, Haozhi},
  booktitle={ICRA},
  year={2026}
}

@article{yin2026osmo,
  title={{OSMO}: Open-source tactile glove for human-to-robot skill transfer},
  author={Yin, Jessica and Qi, Haozhi and Wi, Youngsun and Kundu, Sayantan and Lambeta, Mike and Yang, William and Wang, Changhao and Wu, Tingfan and Malik, Jitendra and Hellebrekers, Tess},
  journal={RA-L},
  year={2026}
}

@inproceedings{he2026forceband,
  title={{ForceBand}: Learning forceful manipulation with {sEMG}},
  author={He, Botao and Wang, Zhi and Kuang, Linna and Ghosh, Ishaan and Malik, Jitendra and Fermuller, Cornelia and Wu, Tingfan and Mao, Jiayuan and Liu, Ruoshi and Qi, Haozhi and Aloimonos, Yiannis},
  booktitle={CoRL},
  year={2026}
}

\end{document}